\pdfoutput=1

\documentclass[letterpaper, 10 pt, conference]{ieeeconf}  % Comment this line out if you need a4paper

\IEEEoverridecommandlockouts                              % This command is only needed if 
\usepackage[utf8]{inputenc} % allow utf-8 input
\usepackage[T1]{fontenc}    % use 8-bit T1 fonts
\usepackage{url}            % simple URL typesetting
\usepackage{booktabs}       % professional-quality tables
\usepackage{amsfonts}       % blackboard math symbols
\usepackage{nicefrac}       % compact symbols for 1/2, etc.
\usepackage{microtype}      % microtypography
\usepackage{lipsum}		% Can be removed after putting your text content
\usepackage{graphicx}
\usepackage{doi}
\usepackage{indentfirst} 
\usepackage{smartdiagram}
\usepackage{subfigure}
\usepackage{multirow}
\usepackage{longtable}
\usepackage{pdflscape}
\usepackage{bbding}
\usepackage{makecell}
\usepackage{longtable}
\usepackage[figuresright]{rotating}
\usepackage{color,xcolor}
\usepackage{algpseudocode}  
\usepackage{amsmath}  
 \usepackage{flushend}

\usepackage{array,graphicx,subfigure}

\usepackage{subfloat}

\usepackage{hyperref} % Latex??url?????????
\usepackage{cite}

\usepackage{threeparttable}

\usepackage{siunitx}
\usepackage{amssymb}

\usepackage{soul, color, xcolor}

\usepackage{stfloats}

\usepackage{changes} %??
\definechangesauthor[name={Jianqi Gao}, color=red]{1} 
\definechangesauthor[name={Jianqi Gao}, color=blue]{2} %

\usepackage[ruled,linesnumbered]{algorithm2e}

\title{RDE: A Hybrid Policy Framework for Multi-Agent Path Finding Problem
}

\definecolor{lime}{HTML}{A6CE39}
\DeclareRobustCommand{\orcidicon}{%
	\begin{tikzpicture}
	\draw[lime, fill=lime] (0,0) 
	circle [radius=0.16] 
	node[white] {{\fontfamily{qag}\selectfont \tiny ID}};
	\draw[white, fill=white] (-0.0625,0.095) 
	circle [radius=0.007];
	\end{tikzpicture}
	\hspace{-2mm}
}

\foreach \x in {A, ..., Z}{%
	\expandafter\xdef\csname orcid\x\endcsname{\noexpand\href{https://orcid.org/\csname orcidauthor\x\endcsname}{\noexpand\orcidicon}}
}
\author{Jianqi Gao\orcidA{}, Yanjie Li\orcidB{},~\IEEEmembership{~Member,~IEEE,} 
	Xiaoqing Yang\orcidC{}, and Mingshan Tan\orcidD{}% 
\thanks{All authors	are with the Department of Control Science and Engineering, Harbin Institute of Technology (Shenzhen), Shenzhen 518055, China (e-mail: gaojianqi205a@stu.hit.edu.cn; autolyj@hit.edu.cn; yxqsheep@gmail.com; tanmingshan033@gmail.com.}% 
}

\begin{document}

\maketitle
\thispagestyle{empty}
\pagestyle{empty}

%%%%%%%%%%%%%%%%%%%%%%%%%%%%%%%%%%%%%%%%%%%%%%%%%%%%%%%%%%%%%%%%%%%%%%%%%%%%%%%%
\begin{abstract}
Multi-agent path finding (MAPF) is an abstract model for the navigation of multiple robots in warehouse automation, where multiple robots plan collision-free paths from the start to goal positions. Reinforcement learning (RL) has been employed to develop partially observable distributed MAPF policies that can be scaled to any number of agents. 
However, RL-based MAPF policies often get agents stuck in deadlock due to warehouse automation's dense and structured obstacles.
This paper proposes a novel hybrid MAPF policy, RDE, based on switching among the RL-based MAPF policy, the Distance heat map (DHM)-based policy and the Escape policy. 
The RL-based policy is used for coordination among agents.
In contrast, when no other agents are in the agent's field of view, it can get the next action by querying the DHM. 
The escape policy that randomly selects valid actions can help agents escape the deadlock.
We conduct simulations on warehouse-like structured grid maps using state-of-the-art RL-based MAPF policies (DHC and DCC), which show that RDE can significantly improve their performance.
\end{abstract}


%%%%%%%%%%%%%%%%%%%%%%%%%%%%%%%%%%%%%%%%%%%%%%%%%%%%%%%%%%%%%%%%%%%%%%%%%%%%%%%%
\section{Introduction}
\label{sec:Introduction}

\begin{figure}[h]%
	\centering
	\includegraphics[width=0.9\linewidth]{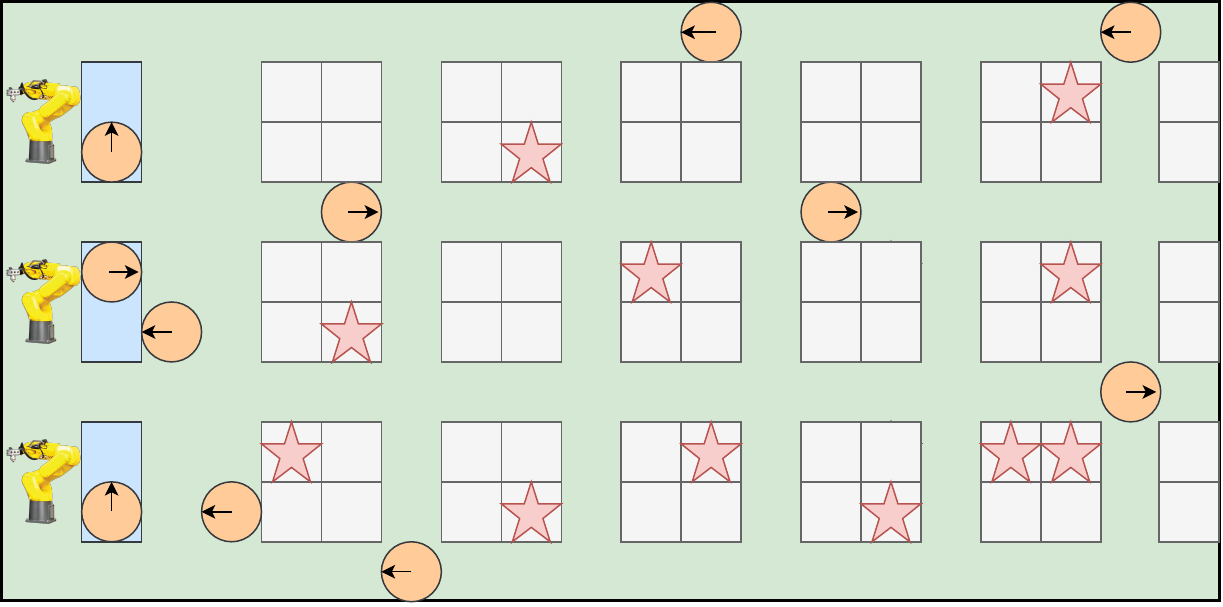}
	\caption{The two-dimensional schematic of the warehouse automation. The orange circles represent the robots. The grey squares represent the inventory pods (obstacles), and the ones marked with light red stars are to be moved. The inventory stations with manipulators are shown on the left side. The robot enters the bottom of the inventory pod, lifts it, and carries it to its designated location.}
	\label{fig:IWS}
\end{figure}

The widespread adoption of e-commerce and logistics has led to the increased use of warehouse automation. As depicted in Fig. \ref{fig:IWS}, a flexible multi-robot system\cite{wurman2008coordinating,berger2015innovative} 
is employed for this purpose. By providing start and goal positions for a group of warehouse robots, MAPF can plan multiple collision-free paths\cite{ma2022graph}. Traditional MAPF policies\cite{ferner2013odrm,sharon2015conflict,cohen2018anytime,silver2005cooperative,barer2014suboptimal,surynek2012towards}
% barer2014suboptimal,surynek2012towards,
have been extensively researched and developed, but their real-time performance deteriorates as the number of agents increases, particularly in large-scale warehouse automation. 

Researchers have recently used RL to acquire distributed MAPF policies, where each agent's action is based on its partial observation space\cite{sartoretti2019primal,liu2020mapper,wang2020,li2020graph,ma2021distributed,ma2021learning}. The RL-based MAPF policy can be applied to any number of agents or map sizes without requiring retraining.
When agents conflict, the RL-based MAPF policy will punish them with negative rewards and make them stay at their current positions. The agent retakes new actions based on the observations in the next step.
However, if the agent's observations do not change in the next step, the RL-based policy will make the agent take the same action as before, leading to a path conflict again. When this process continues, the agent will get stuck in a deadlock, which can lead to the failure of MAPF.
Moreover, the benefits of cooperative coordination using RL-based policy becomes insignificant if the agent has no other agents within its field of view (FOV). Conversely, RL-based MAPF policy may lead to agent detours.

This paper introduces a hybrid MAPF policy called RDE to address the difficulties outlined earlier.
RDE combines the RL-based policy with the DHM-based policy and the escape policy. The RL-based policy mainly manages cooperation and avoids agent conflicts, while the DHM-based policy handles simple scenarios, where no other agents are within an agent's FOV. We can directly query the DHM\footnote{To obtain a DHM with the goal position as the source, we employ the Dijkstra\cite{wang2011application} algorithm to compute the minimum distance to all grid positions. Details can be seen in section \ref{sec:dhm_policy}.} to determine the next action where the agent is closest to the goal position and trying to stay straight.
Additionally, if an agent is trapped in a deadlock state, the escape policy that relies on randomly selecting actions is activated to help resolve the issue. 
To verify the effectiveness of RDE, we utilize state-of-the-art RL-based MAPF policies (DHC\cite{ma2021distributed} and DCC\cite{ma2021learning}) to conduct simulation
experiments on warehouse-like structured grid maps, and the results show that RDE can significantly improve their performance.

The remainder of this paper is structured as follows.
Section \ref{sec:related_work} introduces the research work related to MAPF and lifelong MAPF.
Section \ref{sec:problem formulation} presents the problem formulation. 
Section \ref{sec:Hybrid_plan} describes the hybrid MAPF framework.
%Section \ref{sec:approach} introduces the RL-based policy of RDE.
Section \ref{sec:results} presents the simulation experiment results.
Section \ref{sec:conclusions} provides the conclusions and future work.

\section{Related Work}
\label{sec:related_work}
\subsection{Traditional MAPF}

Finding an optimal MAPF solution has been proven to be NP-hard \cite{yu2013structure}. 
According to whether the optimal solution can be obtained,
we can divide the traditional MAPF policies into optimal and sub-optimal.
The optimal MAPF policies\cite{ferner2013odrm,Sharon2013The,sharon2015conflict,yu2016} also satisfies completeness. 
CBS\cite{sharon2015conflict} is the most commonly used optimal MAPF policy, and researchers have made many improvements to CBS.
However, as the map size or agent's number grows, finding optimal solutions becomes challenging. 
%Many sub-optimal solvers are developed.
A number of sub-optimal policies have been proposed, which can be divided into bounded and unbounded.
%The bounded sub-optimal solvers\cite{cohen2018anytime} can give some guarantee of the quality of the solution. 
The cost of the bounded sub-optimal policy's solution is no more than $(1 + \epsilon) \times c_{opt}$, where $c_{opt}$ is the optimal cost and $\epsilon > 0$ is the sub-optimality factor\cite{barer2014suboptimal}. 
The bounded sub-optimal policies are generally derived from optimal MAPF policies\cite{wagner2015subdimensional,aljalaud2013finding,rahman2022adaptive}. 
Bounded sub-optimal policies can, under certain conditions, improve the speed of solving and provide some guarantee of the quality of the solution.
Unbounded sub-optimal MAPF policies\cite{silver2005cooperative,ma2016multi-agent,Okumura2019,li2021anytime,li2022mapf} can get solutions faster, but the quality of the solution is not guaranteed.

\subsection{Learning-based MAPF}

Learning-based MAPF policies are distributed methods,
%in which each agent only needs the information of FOV to plan a local path.
where every agent takes action based on its partial observation space.
PRIMAL\cite{sartoretti2019primal} combines RL with ODrM$^{*}$\cite{ferner2013odrm}-based imitation learning (IL) to get a distributed MAPF policy.
Based on the framework of PRIMAL, many learning-based MAPF policies are proposed.
MAPPER \cite{liu2020mapper} and G2RL \cite{wang2020} use A$^*$-based shortest path to guide the policy learning. Agents who deviate from this path will be penalized. 
However, A$^*$-based shortest paths are not unique and not optimal globally, destabilizing the learning process and the multi-agentt implicit coordination.
Instead of using a CNN\cite{2022CNN} as the encoder of the observed state, \cite{chen2023transformer} proposes a transformer-based\cite{vaswani2017attention} policy network for feature extraction.
In addition, in the above policies each agent regards other agents as dynamic obstacles, leading to the environment's instability and making the training process challenging to converge. 
Researchers have recently 
%devised a remedy to the issues above by explicitly 
enabled one agent to communicate with other agents.
% within its FOV.
In \cite{li2020graph}, a graph neural network is used to help agents communicate with each other.
Based on\cite{li2020graph} and MAPPER\cite{liu2020mapper}, MAGAT\cite{li2021magat} and AB-Mapper\cite{guan2022ab-mapper} make use of the attention mechanism to assess the relative importance of agent.
DHC \cite{ma2021distributed} uses graph convolution mechanism to communicate between agents.
Instead of broadcast communication in FOV, DCC\cite{ma2021learning} first evaluates the importance of information and then selects some information to aggregate.

%\lipsum[2]

\subsection{Hybrid MAPF Solver}

One policies cannot effectively address every variant of the classical MAPF problem. 
Some researchers attempt to combine multiple MAPF policies to solve MAPF problems.
In \cite{li2021scalable}, the software is proposed to plan paths for thousands of trains within a few minutes, which incorporates many state-of-the-art MAPF policies.
SWARM-MAPF\cite{li2020moving} uses swarm-based policy\cite{jain2016achieving} in open areas and CBS\cite{sharon2015conflict} in areas with dense obstacles, which improves the efficiency of solving problems.
The spatially distributed multi-agent planner \cite{wilt2014spatially} identifies high and low-contention areas and uses different MAPF policies for them.
Some studies use hierarchical policies to solve MAPF problems. First, the map is spatially divided into small areas. The high-level performs global planning for each agent, and the low-level solves the MAPF of each small area.
HMAPP\cite{zhang2021hierarchical} generates the high-level plan for each agent from a randomly picked shortest path from its start position to its goal position and then uses ECBS\cite{barer2014suboptimal} to find conflict-free subpaths in every region.

%\lipsum[2]

\section{Problem Formulation}
\label{sec:problem formulation}

\subsection{MAPF}

The input of MAPF includes an undirected connected graph $G(V,E)$ and an agent set $N=\left \{ 1,...,i,...,m \right \} $. 
Every agent $i$ has a start vertex, $v_i^s \in V$ and a unique goal vertex, $v_i^g \in V$.  
At each discrete time step $t=0,...,\infty $, agent $i$ is located in vertex $v$ and takes action $a_i^t$. Then agent $i$ moves to an adjacent vertex $v^{\prime}$ that meets $(v,v^{\prime}) \in E$  or stays in its current vertex $v$. The action can be represented as $a: v \rightarrow v^{\prime}$ or $a(v)=v^{\prime}$. 
The path of agent $i$ can be represented by a sequence of actions $l_i =(a_i^1,a_i^2,...,a_i^t)$. 
A MAPF solution can be represented as $\mathbb{L} =(l_1,l_2,...,l_m)$.
We assume all agents simultaneously move one grid or stand still at each time step.
In this paper, the world is a two-dimensional four-connected unit grid map where every vertex has four adjacent vertices. The path cost of agent $i$ is the discrete-time $t_i^g$ when the agent $i$ reaches the goal vertex $v_i^g$.
Conflict is usually used in MAPF to represent the path plan collision of different agents. The main conflicts in MAPF are edge conflict $\left \langle i,j,v,v^{\prime},t \right \rangle $ and vertex conflict $\left \langle i,j,v,t \right \rangle $.

\subsection{Learning-based MAPF}

\begin{figure}[ht]
	\centering
	\includegraphics[width=0.8\linewidth]{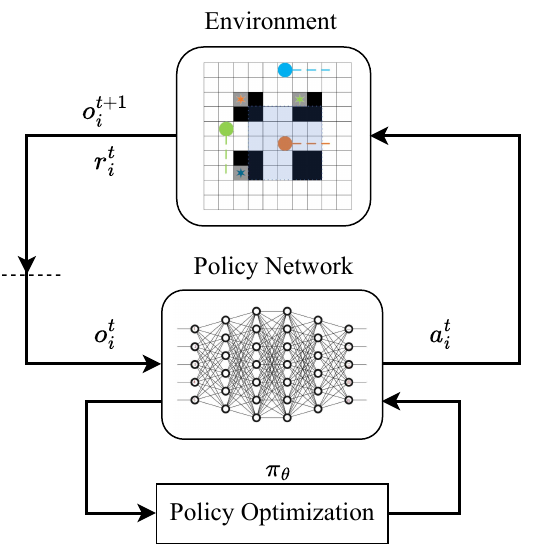}
	\caption{An overview of RL-based MAPF policy. 
		%		MAPF is treated as a partially observable Markov decision process. 
		At each time step $t$, agent $i$ receives its partial observation space ${o}^t_i$ from the environment, which is denoted by ${o}^t_i \in \mathbb{R} ^{W_{FOV}\times H_{FOV}}$, where $W_{FOV}$ and $H_{FOV}$ represent the width and height of the FOV. Then an action $a^t_i$ is generated by the shared policy $\pi_{\theta}$. When the agent reaches the new position in the next time step $t+1$, it will receive a reward $r^t_i$ and its new observation space ${o}^{t+1}_i$. We repeat the above process until the agent reaches the goal position.
	}
	\label{fig:overview_approach}
\end{figure}

As shown in Fig. \ref{fig:overview_approach}, learning-based MAPF can be treated as a partially observable Markov decision process (POMDPs)\footnote{Formally, a POMDP can be represented as $\left\langle   N,\mathcal{S},\left\lbrace \mathcal{A}_i \right\rbrace_{i=1}^N ,\left\lbrace \mathcal{R}_i \right\rbrace_{i=1}^N ,\left\lbrace \mathcal{O}_i \right\rbrace_{i=1}^N ,\mathcal{P},\mathcal{Z},\gamma \right\rangle  $.
	$N$ is the number of agents. $\mathcal{S}$ is the state space containing information about agents and environments. $\mathcal{A}_i$ represents the action space of agent $i$, and $\mathcal{A} = \mathcal{A}_1 \times ... \times \mathcal{A}_n$ is the space of joint actions.
	$\mathcal{R}_i:\mathcal{S}\times \mathcal{A} \rightarrow \mathbb{R}$ is the reward function for agent $i$. $\left\lbrace \mathcal{O}_i \right\rbrace$ is the observation space of agent $i$. $\mathcal{P}(s^{\prime}|s,a):\mathcal{S}\times \mathcal{A}\times \mathcal{S} \rightarrow [0,1]$ is the state transition function, which is used to describe the probability that agent in state $s$ takes action $a$ and then transits to state $s^{\prime}$. $\mathcal{Z}:\mathcal{S}\times \mathcal{A}\rightarrow \mathcal{O}_i$ is the observation function.
	$\gamma \in [0,1]$ is the discount factor.
} \cite{littman1994markov} solved with RL. 
At each timestep $t$, the $i$th agent has an observation $o^t_i$, which only provides partial information of $G(V,E)$, and then we independently compute an action $a^t$ by the policy $\pi_{\theta}$ shared by all agents:
\begin{equation}
{a}^{t}_i \sim \pi_{\theta}\left( {o}^{t}_i\right),  \label{e:policy}
\end{equation}	
where $\theta$ denotes the policy parameters. To find the policy $\pi_{\theta}$, we minimize the expectation of the total path cost of all agents, which is defined as:
%\begin{equation}
%\underset{\pi_{\theta}}{\arg \min } \mathbb{E}\left[ \sum_{i=1}^{m} \lvert l_{i} \rvert \mid \pi_{\theta}\right], \label{e:policy_objective}
%\end{equation}
\begin{equation}
\underset{\pi_{\theta}}{\arg \min } \mathbb{E}\left[ \max_{1 \le i \le m} \lvert l_{i} \rvert \mid \pi_{\theta}\right], \label{e:policy_objective}
\end{equation}
where $\lvert l_{i} \rvert$ is path cost of agent $i$.

\section{Hybrid MAPF Policy}
\label{sec:Hybrid_plan}

%As shown in Fig. \ref{fig:Hybrid_plan_framework}, this section introduces the hybrid MAPF policy framework, which combines the RL-based policy $\pi_{\theta}$ with the DHM-based policy $\pi_{H}$ and the escape policy $\pi_{s}$.

\begin{figure}[htbp]
	\centering
	\includegraphics[width=\linewidth]{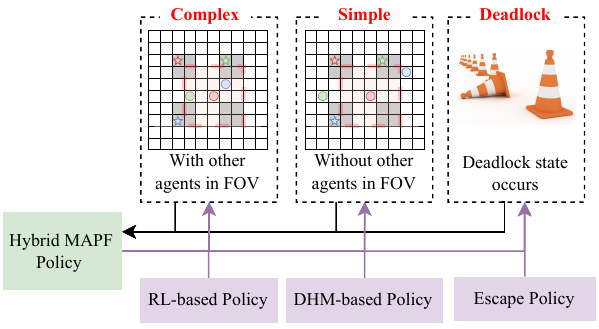}
	\caption{Overview of RDE. The RL-based policy is utilized in complex scenarios $\mathcal{T}_c$, while the DHM-based policy is used for simple scenarios $\mathcal{T}_s$. The escape policy is employed in deadlock state scenarios $\mathcal{T}_d$.
	}
	\label{fig:Hybrid_plan_framework}
\end{figure}

This section presents the hybrid MAPF policy, RDE, depicted in Fig. \ref{fig:Hybrid_plan_framework}. RDE flawlessly combines the RL-based policy $\pi_{\theta}$ with the DHM-based policy $\pi_{H}$ and the escape policy $\pi_{s}$.

\subsection{Classification of Sceniarios}

\subsubsection{Complex}

% 当智能体视野范围内存在其他智能体的时候，智能体不仅关注与障碍物的碰撞，还要考虑与其他智能体之间的协作。我们把这种场景称为复杂。
When there are other agents in the field of view of the agent, the agent not only pays attention to the collision with obstacles, but also considers the cooperation with other agents. We call this scenario as \textit{complex} $\mathcal{T}_c$.

\subsubsection{Simple}

% 当智能体视野范围内没有其他智能体的时候，智能体只关注与障碍物的碰撞，而不需要考虑与其他智能体之间的协作。我们把这种场景称为简单。
When there are no other agents in the FOV of the agent, the agent only pays attention to the collision with obstacles. We call this scenario \textit{simple} $\mathcal{T}_s$.

\subsubsection{Deadlock}
\label{sec:Anomaly_State}

In warehouse automation, encountering deadlock states $\mathcal{T}_d$ like stagnation and oscillation is possible due to the obstacles' high density and structured features\cite{xu2021mapf}. These deadlock states can ultimately fail MAPF.

\textit{Stagnation}.
	If the agent stays at a non-goal position for over $n$ steps, we consider the agent in a state of stagnation.
	When other agents block the warehouse aisle, stagnation happens as the agent cannot find a path.
	We test different values for $n$ and find that $n = 4$ can improve performance.
	
\begin{figure}[h]
	\centering
	\subfigure[$t=1$]
	{
		\label{fig:wander-1}
		\includegraphics[width=0.29\linewidth]{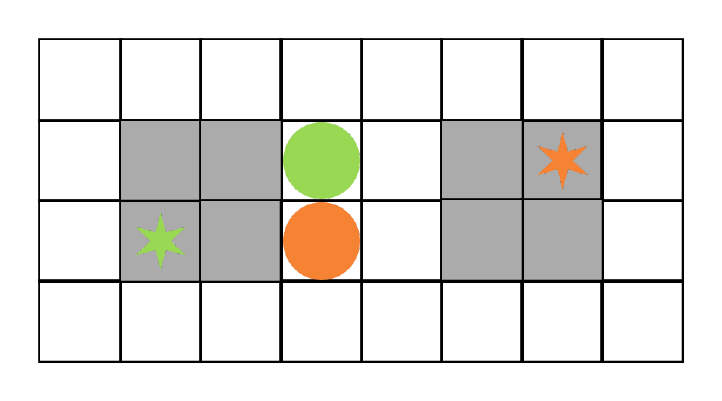}
	}
	\subfigure[$t=2$]
	{
		\label{fig:wander-2}
		\includegraphics[width=0.29\linewidth]{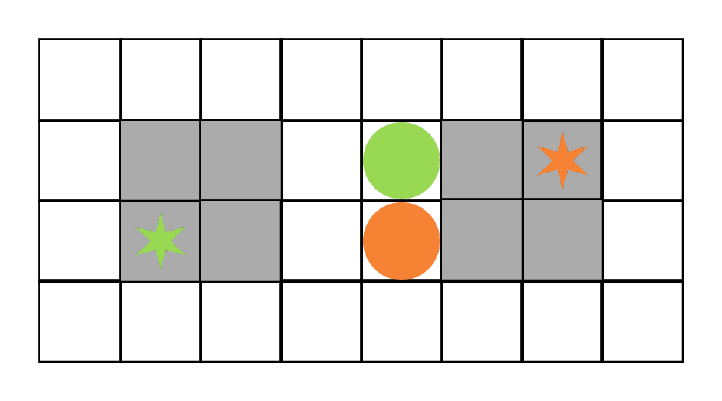}
	}
	\subfigure[$t=3$]
	{
		\label{fig:wander-3}
		\includegraphics[width=0.29\linewidth]{image_wander-3.pdf}
	}
	\caption{An example of the agent in a state of oscillation.
	}
	\label{fig:Occasion_state}
	\vspace{-1em}
\end{figure}

\textit{Oscillation}.
	When a collision occurs between two agents, they may take the same action to avoid it, causing a collision again in the next step. If this process continues, neither agent can reach the goal position. We call this situation oscillation.
	As illustrated in Fig. \ref{fig:Occasion_state}, the green agent moves to the right grid first when time $t=1$, followed by the orange agent. When time $t=2$ arrives, both agents move to the left grid based on their partial observations. In the next time step $t=3$, both agents return to the right grid. 
The single-step oscillation shown in Fig. \ref{fig:Occasion_state} is the most common in the aisle of the warehouse and has the most significant impact on MAPF. Other larger oscillations occur rarely, so we do not consider them in this paper.
% 如何表示振荡
% 其他远距离振荡如何解释

\subsection{RL-based Policy}
\label{sec:RL-based_Policy}

RDE can be adapted to any RL-based MAPF policy.
The state-of-the-art RL-based MAPF policies used in this paper are shown in Table \ref{tab:Comparison_RDE_baselines}. 
%\hl{These solvers were previously trained on random maps.}
%DHC\cite{ma2021distributed} and DCC\cite{ma2021learning} perform excellently and use DHM-based heuristic observation channels. making them ideal for combination with RDE.
DHC\cite{ma2021distributed} and DCC\cite{ma2021learning} perform excellently than other MAPF policies and use DHM-based heuristic observation channels.
We can directly leverage the already established DHM to build a DHM-based policy, which makes DHC and DCC ideal for combination with RDE.

\begin{table}[ht]
	\scriptsize
	\centering
	\setlength\tabcolsep{3.5pt}%???
	\caption{Comparison of RDE and baselines.} 
	\label{tab:Comparison_RDE_baselines}
	\begin{tabular}[c]{cccccc}
		\toprule
		%		\hline
		\textbf{Solver} & \textbf{Learning} & \textbf{Comm.} & \textbf{Guidance} & \textbf{Encoder}& \textbf{Train Env.}\\
		\midrule
		DHC\cite{ma2021distributed}	& Dueling DQN\cite{wang2016dueling} & \checkmark & Heuristic	&CNN& Random\\
		DCC\cite{ma2021learning}	& Dueling DQN\cite{wang2016dueling} & \checkmark & Heuristic &CNN & Random\\
		\bottomrule
		%		\hline
	\end{tabular}
\end{table}

\subsection{DHM-Based Policy}
\label{sec:dhm_policy}

%When most of the agents have reached their goal positions, there may be no other agents within its FOV of the agent. For this simple scenario, Rl-based policy loses its implicit collaboration advantage. 
If no other agents exist in an agent's FOV, RL-based MAPF policy $\pi_{\theta}$ loses its inherent collaborative advantage. Additionally, 
%the RL-based policy $\pi_{\theta}$ can easily cause the agent to oscillate in warehouse automation, where obstacles are very similar.
the dense and structured obstacles in warehouse automation often cause agents to get stuck in deadlock using RL-based MAPF policies.

\begin{figure}[htbp]
	\centering
	\includegraphics[width=\linewidth]{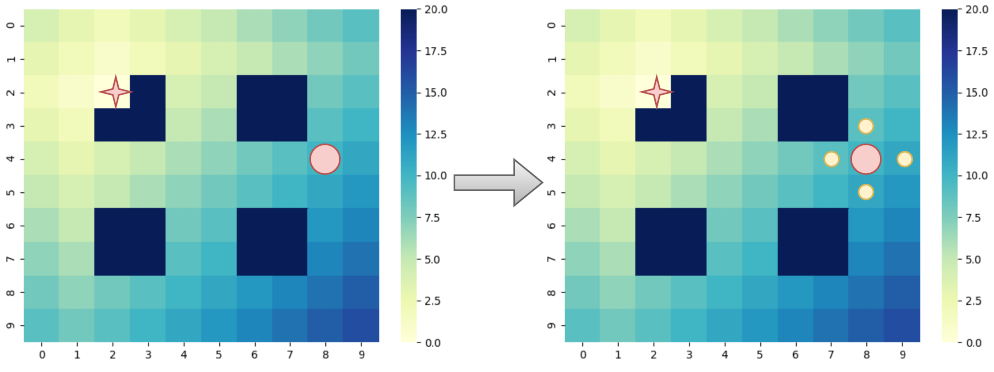}
	\caption{The 10$\times$10 DHM with a star representing the goal position and the dark blue square indicating the obstacles. On the DHM, the heat value of the obstacle position is infinite.
			If the goal position is the same, we can call the existing DHM directly.
%		The distance from the obstacle to the goal position is set to infinity.
	}
	\label{fig:htm}
\end{figure}

The DHM-based policy $\pi_{H}$ is designed in response to the above problems. The policy uses the Dijkstra\cite{wang2011application} algorithm to calculate the shortest distance from the goal position to all grid positions, resulting in a DHM with the goal position as the source. As shown in Fig. \ref{fig:htm}, by querying the DHM, we can determine the shortest distances between the four grids adjacent to the agent's current and goal positions. The position closest to the goal position is then selected as the agent's action $a^{t}_{i,s}$. 

\begin{figure}[htbp]
	\centering
	\includegraphics[width=0.35\linewidth]{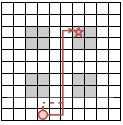}
	\caption{Two paths with the same cost.
		Even though the dashed and solid paths have the same cost, robots that follow the dashed paths will take longer and cause more damage to their hardware in real-world situations. Therefore, reducing the number of turns is essential to optimize its performance and reduce wear and tear.
	}
	\label{fig:samecost}
\end{figure}

Moreover,
when the number of $a^{t}_{i,s} \in \mathcal{A}^t_{i,s}$ is more than one, we prioritize the action that allows the agent to go straight, which satisfies: 
\begin{equation}
\overrightarrow{v^{t-1}_iv^t_i}=\overrightarrow{v^t_iv^{t+1}_i}   , \label{e:straight}
\end{equation} 
where $\overrightarrow{v^{t-1}_iv^t_i}$ represents the current move direction of agent $i$, and $\overrightarrow{v^t_iv^{t+1}_i}$ represents the future move direction of agent $i$. If none of the actions $a^t_{i,s} \in \mathcal{A}^t_{i,s}$ can satisfy Eq. (\ref{e:straight}), the agent randomly selects an action from $\mathcal{A}^t_{i,s}$.
As shown in Fig. \ref{fig:samecost}, 
We try to go straight to reduce robot wear and actual time costs.

\subsection{Escape Policy}

When implementing the RL-based policy in warehouse automation, there is a higher probability for agent to encounter a deadlock state. We introduce an escape policy $\pi_{e}$ to enhance the agent's capability of escaping such a state. As shown in Algorithm \ref{alg:Escape_planner}, if the agent is stuck in a deadlock state while planning, we compel it to randomly choose an action from the available action space, thereby assisting it in breaking free from the deadlock states.

\begin{algorithm}[ht]
	\small
	\caption{The escape policy $\pi_{e}$ for MAPF.}
	\label{alg:Escape_planner}
	\KwIn{Agents set $N=\left \{ 1,...,i,...,m \right \} $, goals set $V^g=\left \{ v_1,...,v_i,...,v_m \right \} $, the position set of each agent for the past five time steps $ V^p_i=\left \{ v^{t-4}_i,v^{t-3}_i,v^{t-2}_i,v^{t-1}_i,v^{t}_i \right \},\forall i\in N $, Action set $\mathcal{A}^{\prime}=\left \{ up, down, left, right \right \}$}
	\KwOut{Actions list: $A=\left \{ a^t_1,...,a^t_i,...,a^t_m \right \}$}
	\For{$i \gets 1$ to $m$}{
		\If{$v^t_i \neq v^g_i$}
		{
			\If{$v^{t-1}_i = v^{t-3}_i \And v^{t-2}_i = v^{t-4}_i$
%				\text{OR} $v^{t-1}_i = v^{t-2}_i \And v^{t-2}_i = v^{t-3}_i$ \label{line:detect}
				}
			{$a^{t}_i \gets random(\mathcal{A}^{\prime})$\; }
		}
	}
\end{algorithm}

\section{Simulation Experiment}
\label{sec:results}

In this section, we show the experiment results of RDE. 
%\hl{Furthermore, we show the validation result of RDE in ROS.}
We implemented the experiments with python.
%Finally, we present the physical robot verification results of RDE.

Simulation is carried out on a single desktop computer with 32 GB memory, Intel$^\circledR$ Core$^\text{TM}$ i7-9700 CPU @ 3.00GHz $\times$ 8 processes and GeForce RTX 2060 SUPER/PCIe/SSE2 graphics. The single desktop computer has a Ubuntu 16.04 LTS system.

\begin{table}[ht]
%	\scriptsize
	\centering
	\setlength\tabcolsep{7.5pt}
	\begin{threeparttable}  
		\caption {Map sets.}
		\label{tab:map_test_agent_density}
		\begin{tabular}[c]{ccccccc}
			\toprule
			\multicolumn{3}{c}{Map}& \multicolumn{4}{c}{Agent Density $\rho_{a}(10^{-2})$\tnote{$^*$}}\\
			\cmidrule(r){1-3} \cmidrule{4-7}
			{Size}&{Type}&{$\rho_{o}$\tnote{$\dag$}} &10&30&50&70\\
			\midrule
			\multirow{2}{*}{$34\times 34$}&\multirow{1}{*}{Sparse}& 0.419 &1.49&4.47&7.44&10.42\\
			&\multirow{1}{*}{ Dense } &0.476&1.65&4.95&8.25&11.56\\
			\bottomrule
		\end{tabular}
		\begin{tablenotes}
			\footnotesize
			\scriptsize
			\item[$\dag$] The density of obstacles $\rho_{o}$ in the map refers to the percentage of staic obstacles.
			\item[$^*$] The agent density $\rho_{a}$ is the proportion of all agents to the vacant positions on the map.
		\end{tablenotes}
	\end{threeparttable}
\end{table}

\subsection{Setup for Testing}

\begin{figure*}[ht]
	\centering
	\subfigure[DHC, Sparse]
	{
		\label{fig:SR_DHC_sparse_small}
		\includegraphics[width=0.225\textwidth]{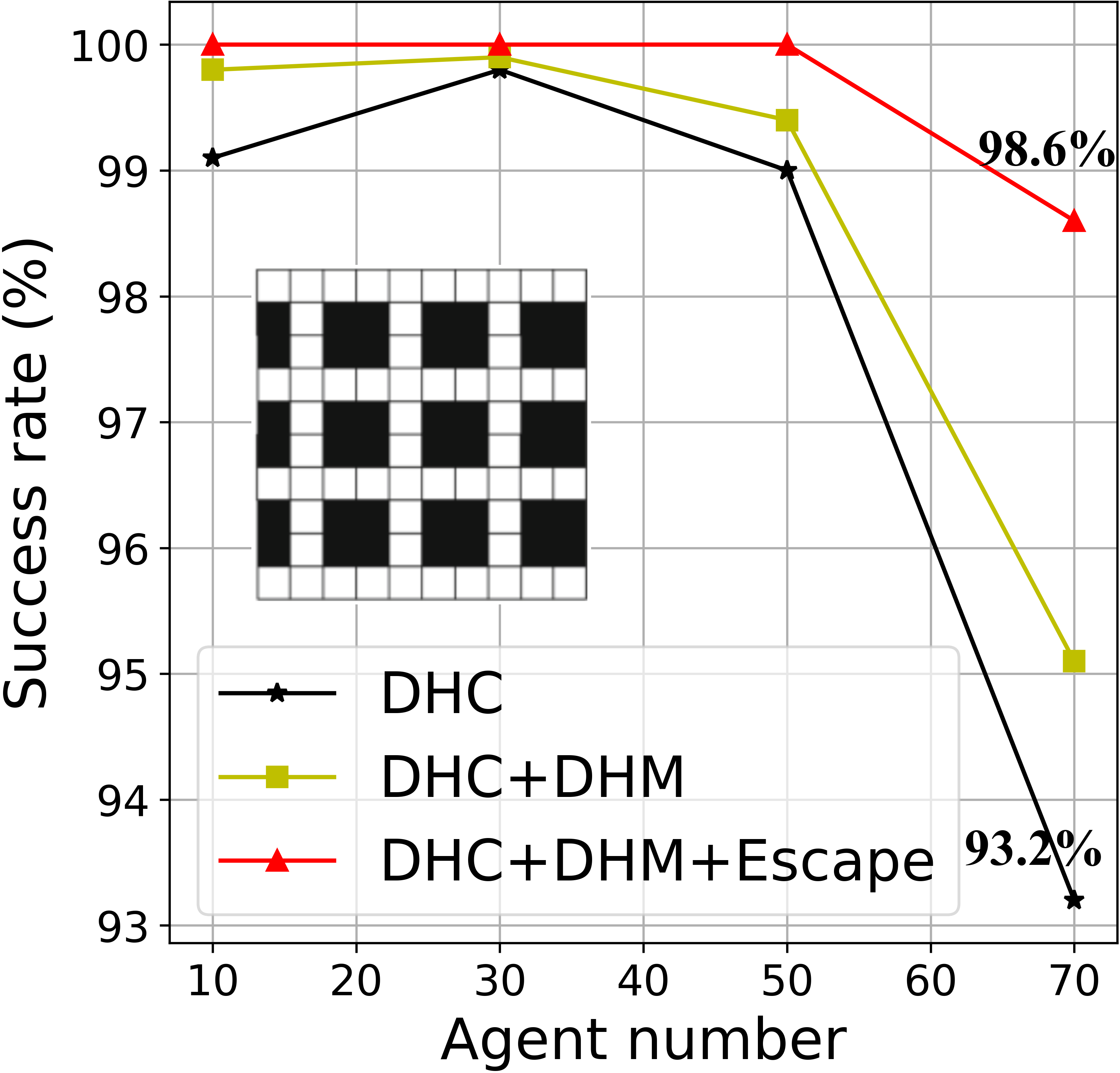}
	}
	\subfigure[DHC, Dense]
	{
		\label{fig:SR_DHC_dense_small}
		\includegraphics[width=0.225\textwidth]{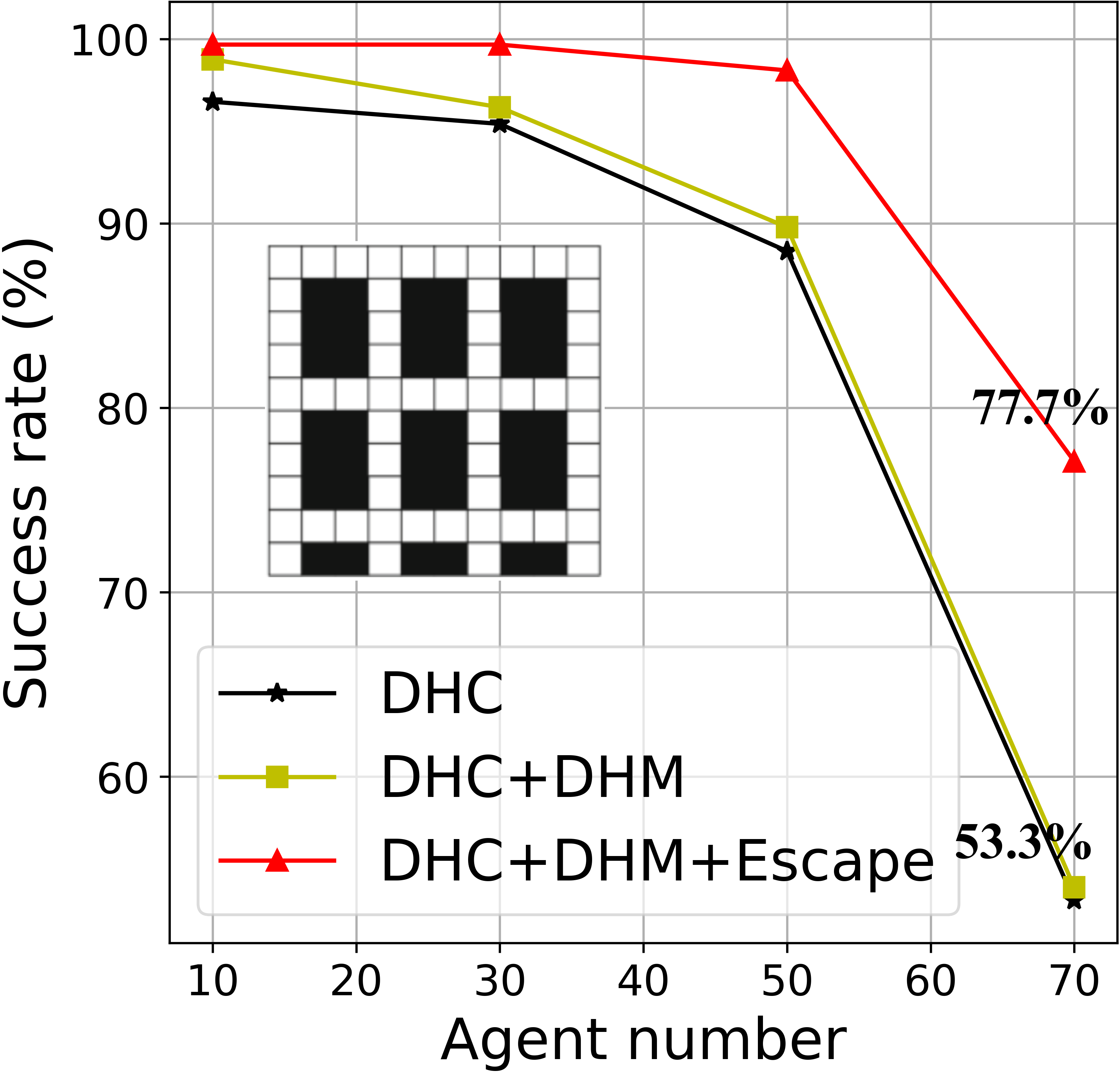}
	}
	\subfigure[DCC, Sparse]
	{
		\label{fig:SR_DCC_sparse_small}
		\includegraphics[width=0.225\textwidth]{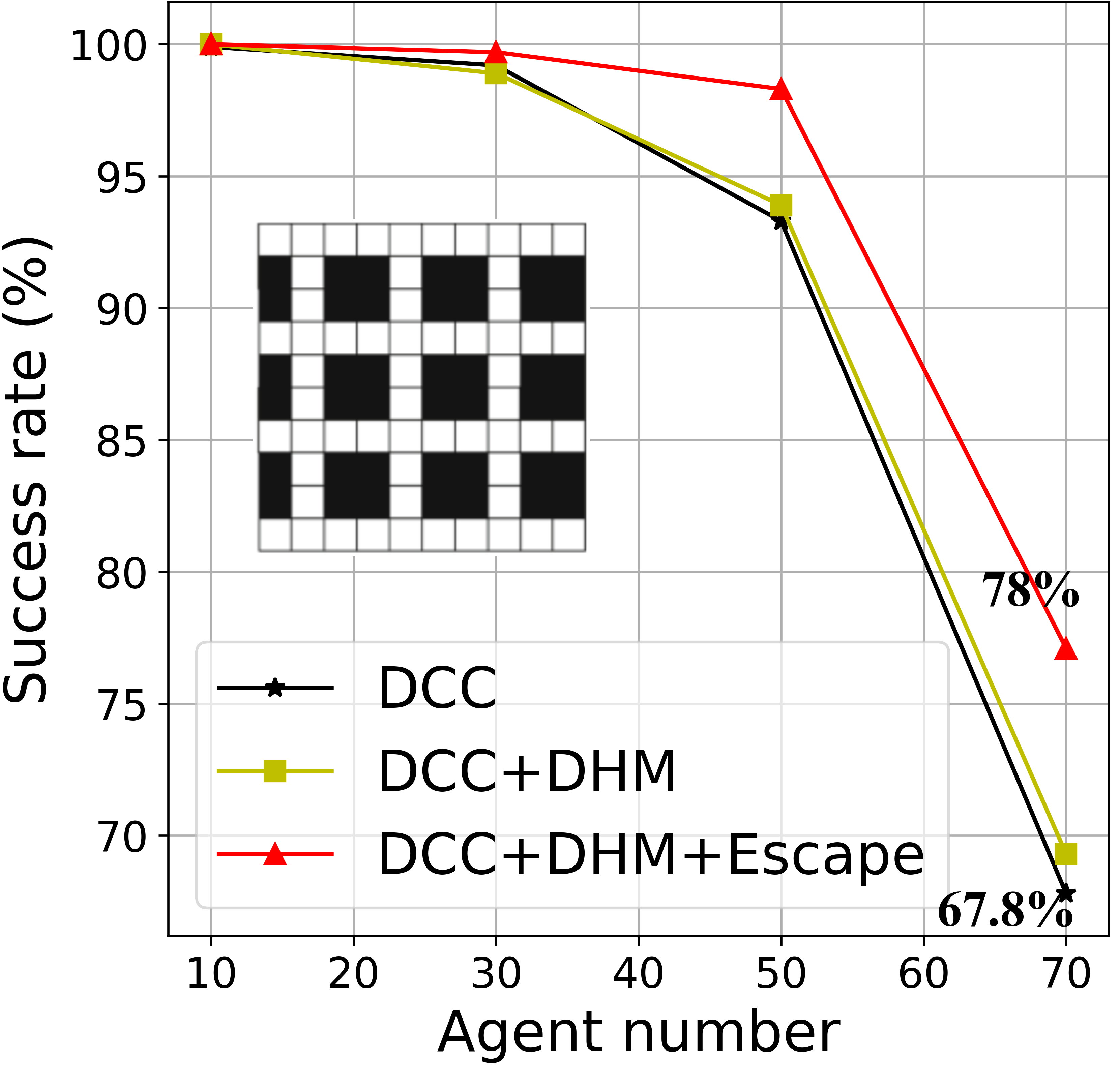}
	}
	\subfigure[DCC, Dense]
	{
		\label{fig:SR_DCC_dense_small}
		\includegraphics[width=0.225\textwidth]{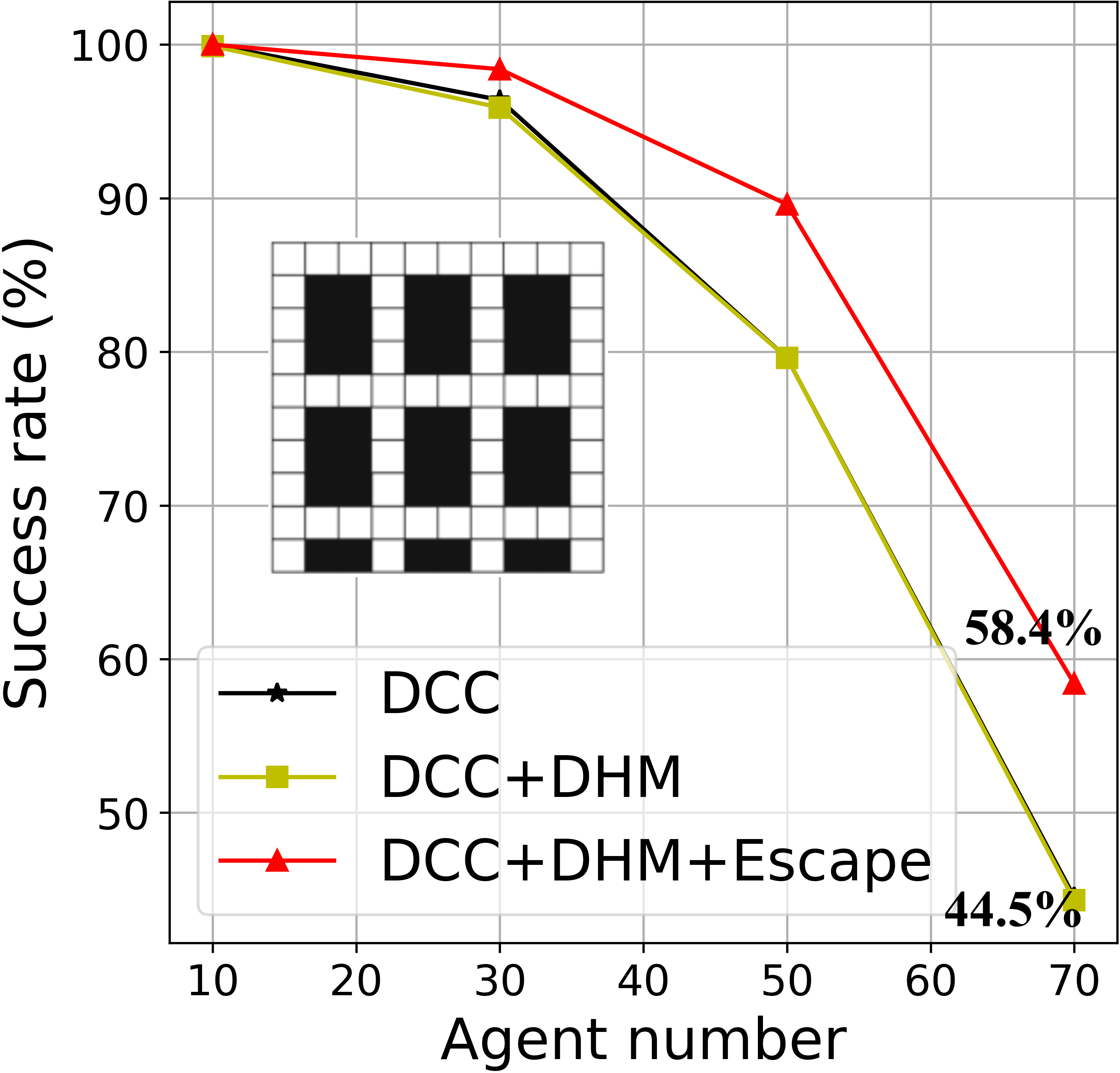}
	}
	\caption{The comparison of $SSR$ for all policies in the small-scale scenarios.}
	\label{fig:SR_small}
	\vspace{-1em}
\end{figure*}

%\subsubsection{MAPF}
%\textit{MAPF}.
As shown in Fig. \ref{fig:SR_small}, 
we choose two kinds of warehouse-like structured maps for testing: \textit{Sparse} and \textit{Dense}. 
Table \ref{tab:map_test_agent_density} shows the obstacle density of two kinds of map are both larger than 0.4, 
much higher than that of other studies\cite{sartoretti2019primal,liu2020mapper,wang2020,li2020graph,ma2021distributed} (up to 0.3). 
There are more path conflicts among agents.
In the small-scale scenarios, the map size is 34$\times$34, and number of agents is 10, 30, 50, and 70, respectively.
We randomly generate 1000 test instances with the same map size and the number of agents for the small-scale scenarios. 
We set the maximum time step size for the small-scale scenario to 150.

As mentioned in Section \ref{sec:RL-based_Policy}, we use state-of-the-art RL-based policies, DHC\cite{ma2021distributed} and DCC\cite{ma2021learning}, to verify the effectiveness of RDE.
We directly use the policy network models of DHC and DCC trained on random maps without retraining on warehouse-like structured maps.
During the test, we combined DHC, DCC with the DHM-based policy (DHC+DHM, DCC+DHM) and then combined DHC, DCC with the DHM-based and escape policies (DHC+DHM+Escape, DCC+DHM+Escape).

\subsection{Metrics}

Success rate ($SSR$). One MAPF instance is deemed unsuccessful when the time step surpasses the maximum value and not all agents reach their goals.  
We usually use the $SSR$ of solving multiple MAPF instances to evaluate the performance of the MAPF policy. $SSR$ can be defined as:
\begin{equation}
\begin{aligned}
SSR = \frac{n_s}{n_t},
\label{e:SR}
\end{aligned}
\end{equation} 
where $n_s$ is the number of successfully solved MAPF instances $\mathcal{I} _s$ and $n_t$ is the total number of instances $\mathcal{I} _t$.

\subsection{Results}

As shown in Fig. \ref{fig:SR_small}, 
RDE can effectively improve the success rate of DHC and DCC on two different  warehouse-like structured grid maps.
As the number of agents increases, the success rate of all policies decreases.
In particular, when the number of agents exceeds 50, the success rates of DHC and DCC drop sharply. By combining DHC, DCC with DHM and escape policies, the sharp decline in success rate has been effectively alleviated.
As shown in Fig. \ref{fig:SR_DHC_sparse_small} and \ref{fig:SR_DHC_dense_small}, when the number of agents is less than 30, the DHM-based policy can further improve the success rate of DHC. Similarly, by comparing Fig. \ref{fig:SR_DHC_sparse_small} with \ref{fig:SR_DHC_dense_small}, and Fig. \ref{fig:SR_DCC_sparse_small} with \ref{fig:SR_DCC_dense_small}, we can also find that the DHM-based policy improves the success rate of the RL-based policy more significantly on sparse maps than on dense maps. It is because when the agent density or obstacle density is low, the possibility of there being no agent in the agent's FOV becomes higher. The DHM-based policy can help the agent take optimal actions to approach the target position.
From Fig. \ref{fig:SR_small}, we can see that the addition of the escape policy has significantly improved the success rates of DHC and DCC on two different maps. Especially on dense maps, when the number of agents reaches 70, the success rates of both DHC and DCC increase by 15\%. Deadlock is the key reason for the failure of MAPF solution, and escape policies based on random action selection can effectively help agents escape from deadlock.

\section{Conclusions and Future Work}
\label{sec:conclusions}

This paper introduces a hybrid MAPF policy framework called RDE, which combines RL-based policy with DHM-based and escape policy for warehouse automation. The RL-based policy can handle cooperative coordination between agents, while the DHM-based policy is suitable for situations where no other agents are within the FOV. In a deadlock, agents can use escape policy to escape from the deadlock.
RDE can be adapted to any RL-based MAPF policy. This paper combines RDE with state-of-the-art RL-based MAPF policies, DHC and DCC. Simulation results show that RDE can further improve the success rate of DHC and DCC.
In RDE, we use an event-based heuristic method for policy switching. However, it may be challenging to cope with more complex scenarios.
In the future, we will seek more intelligent ways to switch policies.

%References are important to the reader; therefore, each citation must be complete and correct. If at all possible, references should be commonly available publications.

\bibliographystyle{IEEEtran}
%\balance
\bibliography{bibliography}

\begin{thebibliography}{10}
\providecommand{\url}[1]{#1}
\csname url@samestyle\endcsname
\providecommand{\newblock}{\relax}
\providecommand{\bibinfo}[2]{#2}
\providecommand{\BIBentrySTDinterwordspacing}{\spaceskip=0pt\relax}
\providecommand{\BIBentryALTinterwordstretchfactor}{4}
\providecommand{\BIBentryALTinterwordspacing}{\spaceskip=\fontdimen2\font plus
\BIBentryALTinterwordstretchfactor\fontdimen3\font minus
  \fontdimen4\font\relax}
\providecommand{\BIBforeignlanguage}[2]{{%
\expandafter\ifx\csname l@#1\endcsname\relax
\typeout{** WARNING: IEEEtran.bst: No hyphenation pattern has been}%
\typeout{** loaded for the language `#1'. Using the pattern for}%
\typeout{** the default language instead.}%
\else
\language=\csname l@#1\endcsname
\fi
#2}}
\providecommand{\BIBdecl}{\relax}
\BIBdecl

\bibitem{wurman2008coordinating}
P.~R. Wurman, R.~D'Andrea, and M.~Mountz, ``Coordinating hundreds of
  cooperative, autonomous vehicles in warehouses,'' \emph{AI Mag.}, vol.~29,
  no.~1, pp. 9--9, 2008.

\bibitem{berger2015innovative}
J.~Berger and N.~Lo, ``An innovative multi-agent search-and-rescue path
  planning approach,'' \emph{Computers \& Operations Research}, vol.~53, pp.
  24--31, 2015.

\bibitem{ma2022graph}
H.~Ma, ``Graph-based multi-robot path finding and planning,'' \emph{Current
  Robot. Reports}, vol.~3, pp. 77--84, 2022.

\bibitem{ferner2013odrm}
C.~Ferner, G.~Wagner, and H.~Choset, ``{OD}r{M$^*$} optimal multirobot path
  planning in low dimensional search spaces,'' in \emph{Proc. IEEE Int. Conf.
  Robot. Autom.}, 2013, pp. 3854--3859.

\bibitem{sharon2015conflict}
G.~Sharon, R.~Stern, A.~Felner, and N.~R. Sturtevant, ``Conflict-based search
  for optimal multi-agent pathfinding,'' \emph{Artif. Intell.}, vol. 219, pp.
  40--66, 2015.

\bibitem{cohen2018anytime}
L.~Cohen, M.~Greco, H.~Ma, C.~Hernandez, A.~Felner, T.~S. Kumar, and S.~Koenig,
  ``Anytime focal search with applications,'' in \emph{Proc. Int. Joint Conf.
  Artif. Intell.}, 2018, pp. 1434--1441.

\bibitem{silver2005cooperative}
D.~Silver, ``Cooperative pathfinding,'' in \emph{Proc. AAAI Conf. Artif.
  Intell. Interact. Digit. Entertain.}, 2005, pp. 117--122.

\bibitem{barer2014suboptimal}
M.~Barer, G.~Sharon, R.~Stern, and A.~Felner, ``Suboptimal variants of the
  conflict-based search algorithm for the multi-agent pathfinding problem,'' in
  \emph{Proc. Int. Symp. Comb. Search}, 2014, pp. 19--27.

\bibitem{surynek2012towards}
P.~Surynek, ``Towards optimal cooperative path planning in hard setups through
  satisfiability solving,'' in \emph{Pacific Rim Int. Conf. Artif. Intell.},
  2012, pp. 564--576.

\bibitem{sartoretti2019primal}
G.~Sartoretti, J.~Kerr, Y.~Shi, G.~Wagner, T.~S. Kumar, S.~Koenig, and
  H.~Choset, ``{PRIMAL}: Pathfinding via reinforcement and imitation
  multi-agent learning,'' \emph{IEEE Robot. Autom. Lett.}, vol.~4, no.~3, pp.
  2378--2385, 2019.

\bibitem{liu2020mapper}
Z.~Liu, B.~Chen, H.~Zhou, G.~Koushik, M.~Hebert, and D.~Zhao, ``{MAPPER}:
  Multi-agent path planning with evolutionary reinforcement learning in mixed
  dynamic environments,'' in \emph{Proc. IEEE/RSJ Int. Conf. Intell. Robots
  Syst.}, 2020, pp. 11\,748--11\,754.

\bibitem{wang2020}
B.~Wang, Z.~Liu, Q.~Li, and A.~Prorok, ``Mobile robot path planning in dynamic
  environments through globally guided reinforcement learning,'' \emph{IEEE
  Robot. Autom. Lett.}, vol.~5, no.~4, pp. 6932--6939, 2020.

\bibitem{li2020graph}
Q.~Li, F.~Gama, A.~Ribeiro, and A.~Prorok, ``Graph neural networks for
  decentralized multi-robot path planning,'' in \emph{Proc. IEEE/RSJ Int. Conf.
  Intell. Robots Syst.}, 2020, pp. 11\,785--11\,792.

\bibitem{ma2021distributed}
Z.~Ma, Y.~Luo, and H.~Ma, ``Distributed heuristic multi-agent path finding with
  communication,'' in \emph{Proc. IEEE Int. Conf. Robot. Autom.}, 2021, pp.
  8699--8705.

\bibitem{ma2021learning}
Z.~Ma, Y.~Luo, and J.~Pan, ``Learning selective communication for multi-agent
  path finding,'' \emph{IEEE Robot. Autom. Lett.}, vol.~7, no.~2, pp.
  1455--1462, 2021.

\bibitem{wang2011application}
H.~Wang, Y.~Yu, and Q.~Yuan, ``Application of dijkstra algorithm in robot
  path-planning,'' in \emph{Proc. IEEE Int. Conf. Mech. Automat. Control Eng.},
  2011, pp. 1067--1069.

\bibitem{yu2013structure}
J.~Yu and S.~LaValle, ``Structure and intractability of optimal multi-robot
  path planning on graphs,'' in \emph{Proc. AAAI Conf. Artif. Intell.}, 2013,
  pp. 1443--1449.

\bibitem{Sharon2013The}
G.~Sharon, R.~Stern, M.~Goldenberg, and A.~Felner, ``The increasing cost tree
  search for optimal multi-agent pathfinding,'' \emph{Artif. Intell.}, vol.
  195, pp. 470--495, 2013.

\bibitem{yu2016}
J.~Yu and S.~M. LaValle, ``Optimal multirobot path planning on graphs: Complete
  algorithms and effective heuristics,'' \emph{{IEEE} Trans. Robot.}, vol.~32,
  no.~5, pp. 1163--1177, 2016.

\bibitem{wagner2015subdimensional}
G.~Wagner and H.~Choset, ``Subdimensional expansion for multirobot path
  planning,'' \emph{Artif. Intell.}, vol. 219, pp. 1--24, 2015.

\bibitem{aljalaud2013finding}
F.~Aljalaud and N.~Sturtevant, ``Finding bounded suboptimal multi-agent path
  planning solutions using increasing cost tree search,'' in \emph{Int. Symp.
  Comb. Search, SoCS}, 2013, pp. 203--204.

\bibitem{rahman2022adaptive}
M.~Rahman, M.~A. Alam, M.~M. Islam, I.~Rahman, M.~M. Khan, and T.~Iqbal, ``An
  adaptive agent-specific sub-optimal bounding approach for multi-agent path
  finding,'' \emph{IEEE Access}, vol.~10, pp. 22\,226--22\,237, 2022.

\bibitem{ma2016multi-agent}
H.~Ma, C.~Tovey, G.~Sharon, T.~S. Kumar, and S.~Koenig, ``Multi-agent path
  finding with payload transfers and the package-exchange robot-routing
  problem,'' in \emph{Proc. AAAI Conf. Artif. Intell.}, 2016, pp. 3166--3173.

\bibitem{Okumura2019}
K.~Okumura, M.~Machida, X.~D{\'e}fago, and Y.~Tamura, ``Priority inheritance
  with backtracking for iterative multi-agent path finding,'' in \emph{Proc.
  Int. Joint Conf. Artif. Intell.}, 2019, pp. 535--542.

\bibitem{li2021anytime}
J.~Li, Z.~Chen, D.~Harabor, P.~J. Stuckey, and S.~Koenig, ``Anytime multi-agent
  path finding via large neighborhood search,'' in \emph{Proc. Int. Joint Conf.
  Auton. Agents Multiagent Syst.}, 2021, pp. 1581--1583.

\bibitem{li2022mapf}
------, ``Mapf-lns2: Fast repairing for multi-agent path finding via large
  neighborhood search,'' in \emph{Proc. AAAI Conf. Artif. Intell.}, 2022, pp.
  10\,256--10\,265.

\bibitem{2022CNN}
Z.~Li, F.~Liu, W.~Yang, S.~Peng, and J.~Zhou, ``A survey of convolutional
  neural networks: Analysis, applications, and prospects,'' \emph{IEEE
  Transactions on Neural Networks and Learning Systems}, vol.~33, no.~12, pp.
  6999--7019, 2022.

\bibitem{chen2023transformer}
L.~Chen, Y.~Wang, Z.~Miao, Y.~Mo, M.~Feng, Z.~Zhou, and H.~Wang,
  ``Transformer-based imitative reinforcement learning for multirobot path
  planning,'' \emph{IEEE Trans. Industr. Inform.}, vol.~19, no.~10, pp.
  10\,233--10\,243, 2023.

\bibitem{vaswani2017attention}
A.~Vaswani, N.~Shazeer, N.~Parmar, J.~Uszkoreit, L.~Jones, A.~N. Gomez,
  {\L}.~Kaiser, and I.~Polosukhin, ``Attention is all you need,'' 2017.

\bibitem{li2021magat}
Q.~Li, W.~Lin, Z.~Liu, and A.~Prorok, ``Message-aware graph attention networks
  for large-scale multi-robot path planning,'' \emph{IEEE Robot. Autom. Lett.},
  vol.~6, no.~3, pp. 5533--5540, 2021.

\bibitem{guan2022ab-mapper}
H.~Guan, Y.~Gao, M.~Zhao, Y.~Yang, F.~Deng, and T.~L. Lam, ``Ab-mapper:
  Attention and bicnet based multi-agent path planning for dynamic
  environment,'' in \emph{Proc. IEEE/RSJ Int. Conf. Intell. Robots Syst.},
  2022, pp. 13\,799--13\,806.

\bibitem{li2021scalable}
J.~Li, Z.~Chen, Y.~Zheng, S.-H. Chan, D.~Harabor, P.~J. Stuckey, H.~Ma, and
  S.~Koenig, ``Scalable rail planning and replanning: Winning the 2020 flatland
  challenge,'' in \emph{Proc. Int. Conf. Automated Plan. Sched.}, 2021, pp.
  477--485.

\bibitem{li2020moving}
J.~Li, K.~Sun, H.~Ma, A.~Felner, T.~S. Kumar, and S.~Koenig, ``Moving agents in
  formation in congested environments,'' in \emph{Proc. Int. Joint Conf. Auton.
  Agents Multiagent Syst.}, 2020, pp. 726--734.

\bibitem{jain2016achieving}
A.~Jain, D.~Ghose, and P.~P. Menon, ``Achieving a desired collective centroid
  by a formation of agents moving in a controllable force field,'' in
  \emph{Indian Control Conference}, 2016, pp. 182--187.

\bibitem{wilt2014spatially}
C.~Wilt and A.~Botea, ``Spatially distributed multiagent path planning,'' in
  \emph{Proc. Int. Conf. Automated Plan. Sched., ICAPS}, 2014, pp. 332--340.

\bibitem{zhang2021hierarchical}
H.~Zhang, M.~Yao, Z.~Liu, J.~Li, L.~Terr, S.-H. Chan, T.~S. Kumar, and
  S.~Koenig, ``A hierarchical approach to multi-agent path finding,'' in
  \emph{Int. Symp. Comb. Search, SoCS}, 2021, pp. 209--211.

\bibitem{littman1994markov}
M.~L. Littman, ``Markov games as a framework for multi-agent reinforcement
  learning,'' in \emph{Machine learning proceedings}, 1994, pp. 157--163.

\bibitem{xu2021mapf}
Y.~Xu, Y.~Li, Q.~Liu, J.~Gao, Y.~Liu, and M.~Chen, ``Multi-agent pathfinding
  with local and global guidance,'' in \emph{2021 IEEE International Conference
  on Networking, Sensing and Control (ICNSC)}, vol.~1, 2021, pp. 1--7.

\bibitem{wang2016dueling}
Z.~Wang, T.~Schaul, M.~Hessel, H.~Hasselt, M.~Lanctot, and N.~Freitas,
  ``Dueling network architectures for deep reinforcement learning,'' in
  \emph{International conference on machine learning}, 2016, pp. 1995--2003.

\end{thebibliography}

\end{document}